\documentclass{jfpda}

\usepackage[pdftex]{pstricks}
\usepackage{verbatim}

\usepackage{natbib}

\usepackage[latin1]{inputenc}
\usepackage[T1]{fontenc}

\usepackage{amsmath,amssymb}
\usepackage{graphicx}
\usepackage[tight]{subfigure}

\title{Actor-critic versus direct policy search: a comparison based on sample complexity}
\author{Arnaud de Froissard de Broissia, Olivier Sigaud}
\institute{
Sorbonne Universit\'es, UPMC Univ Paris 06, UMR 7222, F-75005 Paris, France\\ 
CNRS, Institut des Syst\`emes Intelligents et de Robotique UMR7222, Paris, France\\
{\tt olivier.sigaud@isir.upmc.fr}~~~~+33 (0) 1 44 27 88 53
}

\begin{document}

\maketitle

\begin{abstract}
Sample efficiency is a critical property when optimizing policy parameters for the controller of a robot. In this paper, we evaluate two state-of-the-art policy optimization algorithms. One is a recent deep reinforcement learning method based on an actor-critic algorithm, Deep Deterministic Policy Gradient (DDPG), that has been shown to perform well on various control benchmarks. The other one is a direct policy search method, Covariance Matrix Adaptation Evolution Strategy (CMA-ES), a black-box optimization method that is widely used for robot learning. The algorithms are evaluated on a continuous version of the mountain car benchmark problem, so as to compare their sample complexity. From a preliminary analysis, we expect DDPG to be more sample efficient than CMA-ES, which is confirmed by our experimental results.
\end{abstract}

\section{Introduction}

In order to control more and more mechanically complicated and versatile robotic platforms, robot learning is now a well-accepted necessity in the robotics literature \citep{sigaud10_springer}. The main subfields of robot learning consists either in learning models of the robot or its interactions with the environment (e.g. \cite{salaun10_book,droniou12_iros}), or improving the controller of robots based on experience. In the latter case, the controllers are generally represented as a parametric function of some relevant variables, such as the state of the robot in the case of closed-loop controllers or just a time-related variable in the case of open-loop controllers.
Improving the controller efficiency with respect to some cost function generally requires to perform many evaluations of controllers on the real robot with different parameter values. This process is often time consuming, it may lead to wear and tear of the mechanical structure or even to damage if the tested controllers generate dangerous behaviours. As a result, sample efficiency is a crucial property of any robot learning method.

These methods can be grossly grouped into two main categories:
\begin{itemize}
\item
Direct policy search methods which directly search the space of policy parameters through stochastic optimization, as the name implies.
\item
Actor-critic methods, a subset of reinforcement learning methods \citep{sutton98}, which use an intermediate structure, called the critic, to determine how to update the policy in the direction of greater performance.
\end{itemize}

In \cite{stulp12icml}, the authors showed that, despite an initial attraction towards actor-critic methods such as eNAC \citep{Peters2008NN}, the robot learning literature was converging to black-box optimization methods such as the Covariance Matrix Adaptation Evolution Strategy (CMA-ES), using Dynamic Movement Primitives (DMPs) \citep{ijspeert2013dynamical} as low-dimensional representations.

This temporary supremacy of black-box optimization methods can be explained from several facts. First, actor-critic methods require the approximation of the value or the action-value functions, but the accuracy of this approximation is critical to performance and was limited by the widely used linear function approximators. From one side, simple linear function approximators can be trained with guaranteed convergence, but have a poor representational power, leading to degraded performance. From the other side, more complex, non-linear function approximators can represent more accurately the real value function, but training them cannot be guaranteed to converge \citep{baird93}. Second, real-world robotic control problems may require the use of a large state representation, and all the actor-critic methods cited above did not scale well in that respect until recently.

The situation changed drastically with the recent publication of several ``deep'' reinforcement learning algorithms.
The discrete action Deep Q-Network (DQN) algorithm \citep{mnih2015human} and its continuous action, actor-critic counterpart, Deep Deterministic Policy Gradient (DDPG), \citep{lillicrap2015continuous} overthrew these limitations by making the training process of the value function approximator more stable, robust, and scalable. 
The wide applicability of DDPG to several benchmarks is quite impressive, but the paper was published without a performance comparison with any other method.
Recently, a general comparison of many robot learning methods was published based on several simulation benchmarks \citep{duan2016benchmarking}, but it only compares the final controller performance and does not come with a detailed analysis of why some methods outperform others.

In this paper, we focus on sample efficiency and show that, in addition to begin much more scalable, the actor-critic approach in DDPG is also much more sample efficient than the direct policy search approach of CMA-ES, even for a 280 parameters controller applied to the small-size mountain car benchmark.

The paper is organized as follows. In Section~\ref{sec:algos}, we quickly present both compared algorithms.
In Section~\ref{sec:setup} and \ref{sec:result}, we respectively describe the experimental set-up of the comparison and the corresponding results.
The significance of these results is discussed in Section~\ref{sec:discu}, before we conclude and highlight directions for future work.

\section{Algorithms}
\label{sec:algos}

In this Section, we shortly describe the training mechanisms in DDPG and CMA-ES so as to higlight their differences.

\subsection{Deep Deterministic Policy Gradient}

Deep Deterministic Policy Gradient \citep{lillicrap2015continuous} is an actor-critic algorithm using deep neural networks to represent both the value function and the policy over a continuous state-action space. It combines ideas from DQN \citep{mnih2015human}, Deterministic Policy Gradient (DPG) \citep{silver2014deterministic} and batch normalization \citep{ioffe2015batch}.

In DDPG, the actor network deterministically maps a state vector to an action vector, thus learning a deterministic policy, which is easier than learning a stochastic one, the search space being smaller.

We note hereafter $t$ the current time step, $s_t$ the state vector at $t$, $a_t$ the action vector at $t$ and $r_t$ the reward at $t$.

When interacting with the environment, each $(s_t, a_t, r_t, s_{t+1})$ sample is stored in a replay buffer.
During training, a minibatch of samples is randomly drawn from the replay buffer, making them seemingly i.i.d, which is a key trick borrowed from DQN to improve the stability of the algorithm. 

A second trick is borrowed from DQN to improve stability. Instead of directly using the actor and the critic networks to perform the standard temporal computation, two networks called ``target networks'' are used. These networks ensure more stable computation because they are updated more slowly. In practice, they track the current networks using $\theta' = \theta'\times(1-\tau) + \theta\times \tau$ with $\tau$ small, where $\theta$ is the set of parameters of the considered network.

Thus, the critic is trained to learn the state-action value function by minimizing the temporal difference error using
\[
\delta_t = r_t + \gamma Q'(s_{t+1},\pi'(s_{t+1})|\theta') - Q(s_t,a_t|\theta),
\]
where $\gamma$ is the discount factor, $Q$ is the current critic, $Q'$ is the target critic, $\pi'$ is the target policy and $\theta$ (resp. $\theta'$) are the parameters of the critic (resp. target critic) networks.
The algorithm minimizes the squared error over the minibatch through gradient descent, using the loss function
\[
L = 1/N \sum_{i\in m} {\delta_i}^2,
\]
where $N$ is the size of the minibatch and $m$ the content of the minibatch. 
Thanks to the generalization property of neural networks, the critic performs accurate approximation of the action-value function for each point of the state space without the need for a lot of samples. Note however that gradient descent is a local optimization method and does not have any guarantee to converge to a global optimum. 

Then, the actor is trained using the gradient of the deterministic policy, as proposed in \cite{silver2014deterministic}:
\[
\nabla_w \pi (s,a) = \mathbb{E} \rho (s)[\nabla_a Q(s,a|\theta)\nabla_w \pi (s|w)].
\]

This gradient is calculated by first backpropagating the gradient of the value function with respect to the actions through the critic. Computing the gradient with respect to actions is similar to doing so with respect to weights, as already noted in \citep{hafner2011reinforcement}. Then the algorithm backpropagates the obtained gradient in the actor with respect to its parameters from its output layer to its input layer. The whole training computation thus relies on efficient gradient backpropagation algorithms provided by any deep learning library (here we use TensorFlow).
Actually, the gradient propagated over the actor network expresses in which direction to move in the policy parameter space to get better outcomes for a given state.

The third trick, batch normalization, also improves stability and accelerates learning. Batch normalization is not used in the experiments hereafter, and we will not describe it here. The reader is refered to the original paper \citep{ioffe2015batch} for a description.

The actor and critic networks are trained after each step in the environment. Although the training process inherits off-policy properties from DPG, it is performed in parallel to running an episode, thus the algorithm improves its policy while using it to interact with the environment. However, training of the networks can be more or less decoupled from the sample acquisition process depending on the replay buffer data management policy, which can be critical for the efficiency of the algorithm \citep{debruin15}.

Here we use a replay buffer with a maximum size $M$, where new samples are added until the maximum capacity is reached. From there, for each new sample, a previously stored sample is randomly deleted from the replay buffer. As proposed in \citep{debruin15}, the $F$ first samples are kept and are not replaced with new samples to keep a pool of initial samples coming from non-optimal trajectories. 

\subsection{Covariance Matrix Adaptation Evolution Strategy}

Covariance Matrix Adaptation Evolution Strategy \citep{hansen2003reducing} is a gradient-free evolutionary method, using random variations to improve a set of real-valued parameters relatively to an objective function. The general idea consists in representing a distribution over sets of parameter values through a covariance matrix, evaluating each set of parameters and updating the covariance matrix towards better performance.

In the context of robotics experiments, the objective function is the outcome of one or several episodes of the considered task and the parameters are those of the controller that runs the episodes.


Thus, at each training step, that is after each whole batch of episodes, a population of test controllers is sampled around the current one using the covariance matrix, and evaluated on the task. One can immediately see that performance improvement relies on running many episodes, which is not sample efficient, whereas in actor-critic methods like DDPG, the actor can be updated at each step just from the gradient of the critic, without requiring any new samples, provided that the replay buffer provides good enough information to update the critic.

\section{Experimental set-up}
\label{sec:setup}

Our goal in this paper is to compare DDPG and CMA-ES in terms of sample efficiency. 

A task is characterized by a state space, a transition function that, given the current state and action, gives a probability distribution over the next state, and a reward function that, given a transition, gives a scalar. Here, we restrict our analysis to episodic tasks that have one starting state and potentially several terminal states.

We evaluate the performance of both algorithms on a continuous version of the mountain car benchmark. In this task, a car is placed between two hills, and has to reach a target on the top of one hill. The car does not have sufficient power to reach the reward by driving directly towards the target, and needs to gain momentum by going up and down the slopes of both hills. A wall prevents the car from going to far away from the non rewarded hill. This task is deterministic: given a state-action pair, there is a single corresponding next state. The task ends when the car reaches the top of the hill or the maximum time $T$ has been reached.

The reward signal is a positive scalar $R$ obtained when reaching the target. The use of a discount factor in DDPG favors shorter trajectories to the target. In order to incorporate the same drive towards shorter trajectories in CMA-ES, we discount the final reward in CMA-ES using the same discount factor $\gamma$ as in DDPG.
A cost proportional to the square of the applied action at all steps is also added, using a cost coefficient $\rho$. This signal generates a strong local optimum that corresponds to not moving, allowing to test the exploration efficiency of both algorithms. No negative scalar is received when reaching the maximum amount of time as the state vector does not include time, and the task has to be fully observable.

In order to facilitate the comparison, both algorithms are run on the same controller structure, that is a multi-layer neural network.
Since CMA-ES uses a covariance matrix, its space complexity is quadratic, so the number of actor parameters must be kept low. 
In order to determine the adequate dimension for a network, we started with very small networks and increased the size as long as
CMA-ES performance was improving for a reasonnable computational budget.
In order to make the comparison more fair, we decided not to incorporate batch normalization in DDPG, because using batch normalization in DDPG would require adding several dedicated layers of neurons in the actor network, thus would result in a different structure for the DDPG and the CMA-ES actor networks.

The resulting actor network has 2 units in its input layer (one for the car position, one for its speed), 2 hidden layers with $h_1$ and $h_2$ units respectively and 1 output unit (the positive or negative acceleration, constrained in the range of actions $\delta$ = [-1,1]), for a total of 51 parameters. As in the original article describing DDPG, the first hidden layer uses the rectified non-linearity as unit transfer function and the second the $tanh$ function.
Furthermore, the critic in DDPG contains 2 hidden layers of 20 and 10 units respectively, with the same internal structure(rectified non-linearity and tanh). Actions are added only after the first hidden layer. The learning rate of the critic, $\alpha_c$, is 0.005, that of the actor, $\alpha_a$, is 0.01. After each action step of DDPG, a training step is performed using one minibatch of $N$ samples.

All the meta-parameters of the experimental study are shown in Table~\ref{tab:ddpg}~\footnote{
The source code of the experiment is available online: \textbf{https://github.com/MOCR/DDPG}}.

\begin{table}[hbtp!]
\begin{center}
\begin{tabular}{|c|l|l|l|}
\hline
Entity & Parameter & Value & Meaning\\
\hline
& $M$ & 100000&replay buffer size\\
& $F$ & 20000&first samples kept\\
& $\tau$ &  0.001&target update factor\\
DDPG & $N$ & 64& minibatch size\\
& $h_1$ & 5 & nb neurons in hidden layer 1\\
& $h_2$ & 5 & nb neurons in hidden layer 2\\
& $\alpha_c$ & 0.005 & learning rate of critic\\
& $\alpha_a$ & 0.01 & learning rate of actor\\
& $ \gamma$ & 0.99 & discount factor\\
\hline
CMA-ES & $ \sigma$ & 0.5&variance for exploration\\
\hline
& $ T$ & 999&max nb of time steps\\
mountain & $R$ & 100&reward value\\
car& $\rho$ & $ 0.1 \times a^2$& coefficient of energy cost\\
& $\delta$ & $ [-1 ; 1]$&range of action\\
\hline
\end{tabular}
\caption{Meta-parameters of DDPG, CMA-ES and the benchmark used for the experiments\label{tab:ddpg}}
\end{center}
\end{table}

When solving this task, we are interested in how much data from the environment is needed to learn a good policy for both algorithms. 
Therefore the metric we use is the number of interactions with the environment. 

For each performed action, DDPG goes through one training step on a single batch of samples. But using multiples batchs may provide a faster convergence, and consequently requires less interactions with the environment to get to the same performance. Therefore, we also compare DDPG with one minibatch per training step to DDPG with four minibatches per training step. 

\pagebreak
\section{Simulation results}
\label{sec:result}

All performance curves shown below are averaged over 10 runs on all figures, and are obtained in less than one hour on a small CPU cluster with 16Go RAM nodes cadenced at 2.26 Ghz.

\begin{figure}[hbtp!]
\centering
{
\subfigure[\label{traj_CMA}]{\includegraphics[width=0.48\hsize]{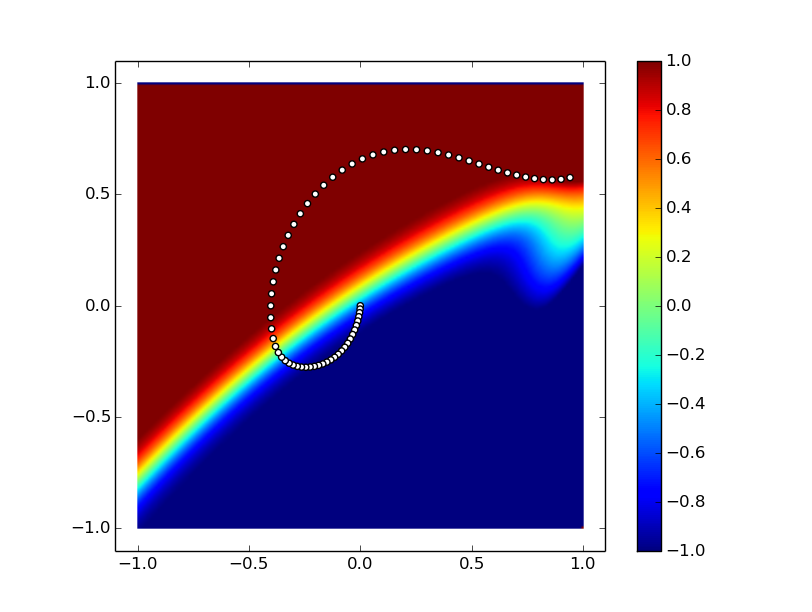}}
\subfigure[\label{traj_DDPG}]{\includegraphics[width=0.48\hsize]{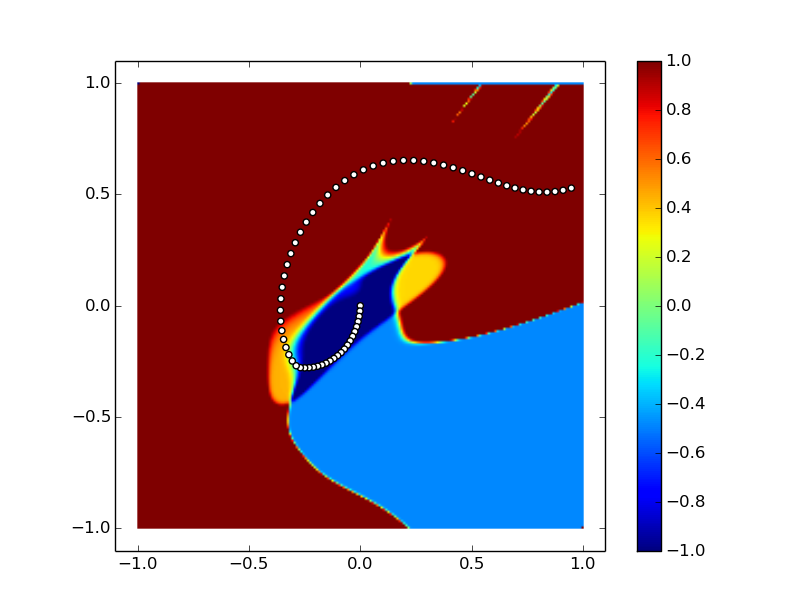}}
}
\caption{Final policies obtained with CMA-ES (a) and DDPG (b). The $x$ and $y$ coordinates correspond to position and velocity, and the color scale to the positive or negative acceleration.\label{fig:ddpgvscma_pol}}
\end{figure}

Figure~\ref{fig:ddpgvscma_pol} illustrates the final policies obtained with CMA-ES and DDPG on the mountain car problem. One can see that the policy found with DDPG shows a better generalization outside the illustrated trajectory than the one found with CMA-ES.

\begin{figure}[hbtp!]
\centering
{
\subfigure[\label{time_CMA}]{\includegraphics[width=0.48\hsize]{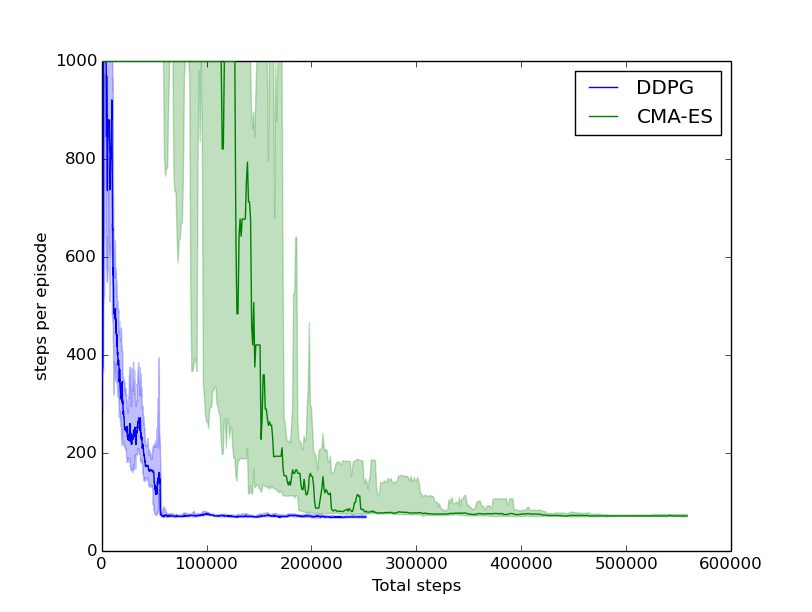}}
\subfigure[\label{time_DDPG}]{\includegraphics[width=0.48\hsize]{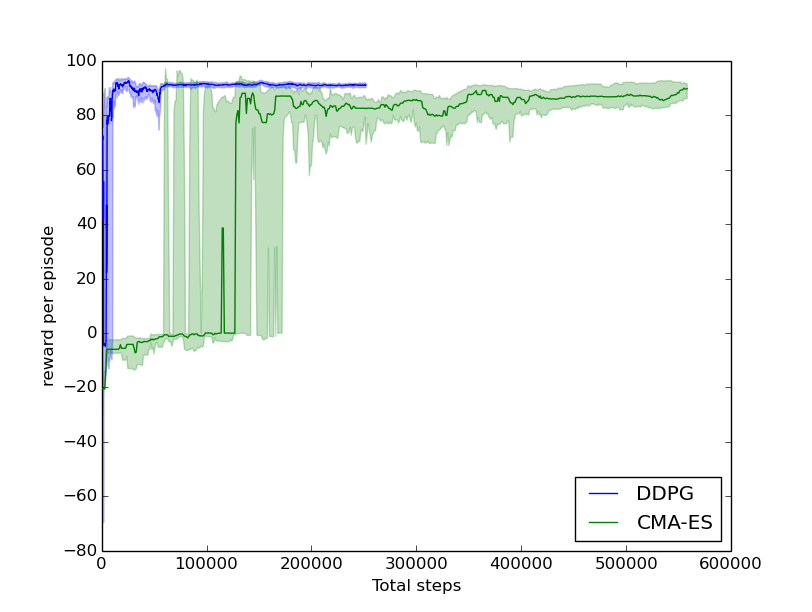}}
}
\caption{Time (a) and reward (b) per episode over total number of interactions with the environment.\label{fig:ddpgvscma_time}}
\end{figure}

Figure~\ref{fig:ddpgvscma_time} shows the evolution of the learning performance in terms of the time needed to reach the target and collected reward with DDPG and CMA-ES.

One can see that in terms of time per episode as well as collected reward, DDPG converges faster in terms of number of interactions, with less variance over different runs, and is significantly more sample efficiency than CMA-ES.

Actually, on Figure~\ref{time_DDPG}, the best performance found over all the CMA-ES evaluations is slighly better than the one found with DDPG, but this does not seem to be significant, DDPG being better on average. By the way, although their experimental settings differ from ours, \cite{duan2016benchmarking} also find better performance with CMA-ES compared to DDPG in some cases.

\pagebreak

\begin{figure}[hbtp!]
\begin{center}
  \includegraphics[width=.6\textwidth]{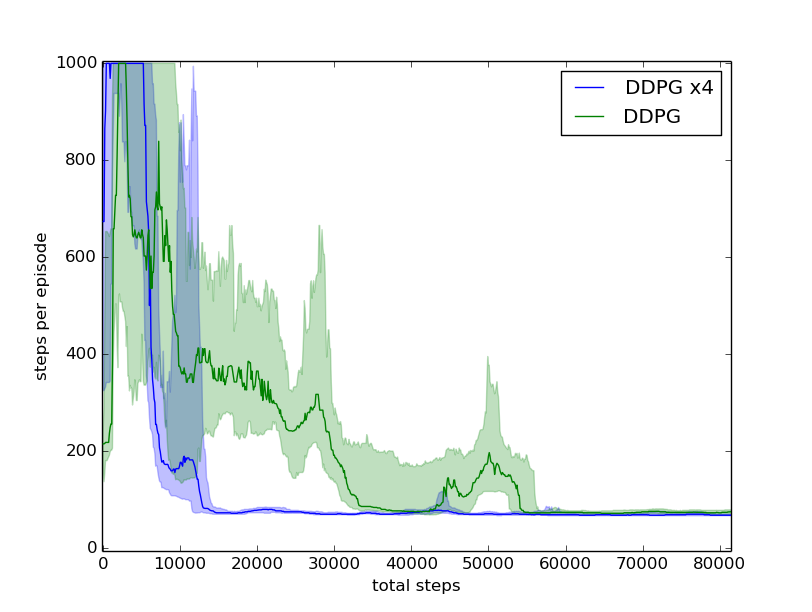}
\caption{Time per episode over total number of interactions with the environment for DDPG with one minibatch per training step and DDPG with four minibatches per training step.\label{fig:ddpgx4}}
\end{center}
\end{figure}

Figure~\ref{fig:ddpgx4} shows the performance of DDPG when performing either one or four minibatches training iterations per training step. 

By using more minibatches between each step in the environment, DDPG requires even fewer interactions with the environment to converge.

\begin{figure}[hbtp!]
\begin{center}
\includegraphics[width=.6\textwidth]{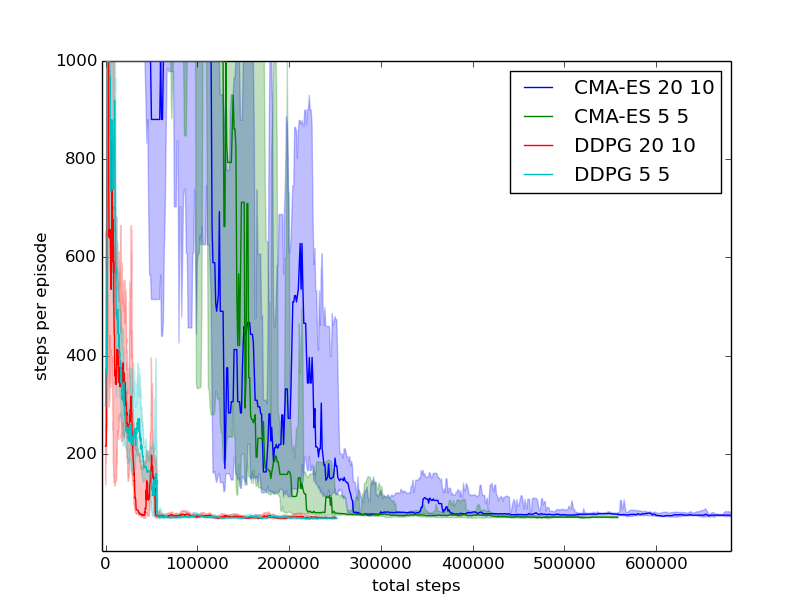}
\caption{Time per episode over total number of interactions with the environment for DDPG and CMA-ES with different sizes of the actor network.\label{fig:size}}
\end{center}
\end{figure}

Finally, Figure~\ref{fig:size} shows the impact of the size of the actor on the learning performance of DDPG and CMA-ES.

Using a larger network with $h_1=20$ and $h_2=10$, resulting in 281 parameters to be optimized, has no impact on the learning performance of DDPG, whereas it is slightly detrimental to the convergence and stability of CMA-ES on the long run. However, with even larger actors, CMA-ES is slower to converge, and quickly runs out of memory for too large sizes, as already pointed out in  \cite{duan2016benchmarking}.

\section{Discussion}
\label{sec:discu}

We have chosen the Mountain car problem because this facilitates the comparison between DDPG and CMA-ES, due to the low dimensionality of the task. It should be noted that using a continuous version of this problem ``makes the task easier for the CMA-ES and more difficult for the NAC'' \citep{heidrich2008variable}, NAC being an actor-critic ancestor of DDPG. The experimental results described above have been obtained with a rather small actor network. 

Based on our evaluations, the general finding is that DDPG requires far less interactions with then environment than CMA-ES and with less variance between different runs. 
In itself, this superiority is not surprising, as it may results from various facts:
\begin{itemize}
\item
{\bf analytic gradient descent versus stochastic gradient-free search}: DDPG relies on optimized analytical gradient descent algorithms provided in deep learning toolboxes, whereas CMA-ES is gradient-free and relies on somewhat blind parameter exploration. However, CMA-ES implements reward-weighting averaging, which has been shown to be an approximate way to perform approximate natural gradient descent \citep{stulp2012policy,stulp13paladyn}. Whether analytic vanilla gradient descent is more efficient than approximate natural gradient descent is an open question that needs to be investigated in the near future.
\item
{\bf better reuse of sample data}: Both algorithms have a very different way of using the environment. 
Whereas DDPG first collects samples and afterward update policy parameters to adapt to what was collected, CMA-ES first stochastically samples new parameters and then evaluates how they perform. The former uses the environment as a source of information and the latter as a source of evaluation. An other difference is that, in DDPG, the collected information stays valid and can be stored into the replay buffer for subsequent training, whereas in CMA-ES training is intrinsically local to a set of parameters, thus the evaluation samples cannot be stored or reused. Those two differences partly explain why DDPG requires less interactions with the environment to converge.
\item
{\bf actor-critic versus direct policy search}:
As clearly explained in \citep{sutton00_NIPS}, a critic is an efficient way to summarize the performance of a system along a trajectory, without having to perform this trajectory. Part of the better sample efficiency of DDPG with respect to CMA-ES certainly comes from the fact that the policy can be improved without calling upon new samples, once the critic correctly approximates the performance of the current policy.
\end{itemize}

\section{Conclusion and future work}

Recent deep reinforcement learning algorithms have opened the way to many new applications due to their unprecedented scalability \citep{duan2016benchmarking}. In this paper, we have disregarded scalability to rather focus on sample efficiency. We have provided a sample efficiency comparison between the training mechanisms of DDPG and CMA-ES on a continuous version of the small mountain car benchmark problem, using deep neural networks as policy representation. Our results indicate that the DDPG mechanisms are significantly more sample efficient than those of CMA-ES. This sample efficiency is likely to reside in the use of a replay buffer, but also in the more efficient gradient descent algorithm.

However, the above comparison is limited in several respects. First, the CMA-ES and DDPG training processes were compared using a neural network as policy representation, but using an open loop controller representation based on DMPs as is often done in robot learning would probably be more favorable to CMA-ES. 
Evaluating DDPG on neural networks versus CMA-ES on DMPs in terms of required samples to converge would be a relevant comparison for robotics that remains to be performed. One may also consider using DDPG on DMPs, but this approach would not make profit of the scaling capability of DDPG while still being subject to the drawbacks of DMPs (see \cite{stulp13paladyn} for a discussion). Second, we have not incorporated some of DDPG mechanisms such as batch normalization. Assessing the influence of such processes might be of interest too. Third, it would be of much interest to disentangle the respective role of the various factors hightlighted in the above discussion to explain the superior sample efficiency of DDPG. This can be done by comparing the performance of impoverished versions of both algorithms where the sources of the various factors are neutralized one by one. This is one of the main items in our agenda for future research.
Finally, the publication of DDPG has drawn attention on deep reinforcement learning as an emerging domain, and several even more recent algorithms such as \citep{heess2015learning,balduzzi2015compatible,heess2015memory,gu2016continuous} also deserve to be studied in terms of their elementary mechanisms and efficiency factors.

\section*{Acknowledgments}
This work was supported by the European Union's Horizon 2020 research and innovation program within the DREAM project under grant agreement No 640891.

\bibliography{deep,perso,rl,continuous_rl}

\end{document}